%% file: main.tex
\documentclass[11pt,onecolumn,letterpaper]{article}
\usepackage[margin=0.8in]{geometry}
\usepackage{mathptmx}
\usepackage{microtype}
\usepackage[numbers,sort&compress]{natbib}
\usepackage[hyphens]{url}
\input{preamble}
\usepackage[font=small,labelsep=period]{caption}
\definecolor{linkblue}{rgb}{0.16,0.36,0.57}
\usepackage[colorlinks,allcolors=linkblue]{hyperref}

\title{TrajRealm: A Large-Scale Point Tracking Dataset with Diverse Dynamics}
\author{%
  Ni Yang\textsuperscript{1,*}\quad
  Weiguang Zhao\textsuperscript{2,*}\quad
  Jiafei Cao\textsuperscript{2}\quad
  Yankai Jiang\textsuperscript{2}\\[2pt]
  Haoyu Guo\textsuperscript{2}\quad
  Tao Yang\textsuperscript{1,\ensuremath{\dagger}}\quad
  Chunhua Shen\textsuperscript{3}\quad
  Junting Dong\textsuperscript{2,\ensuremath{\dagger}}\\[5pt]
  {\small \textsuperscript{1}Fudan University\quad
   \textsuperscript{2}Shanghai AI Laboratory\quad
   \textsuperscript{3}Zhejiang University}\\[3pt]
  {\small \textsuperscript{*}Equal contribution\qquad
   \textsuperscript{\ensuremath{\dagger}}Co-corresponding authors}\\[4pt]
  {\small \textbf{Project page:} \href{https://purplehihi.github.io/TrajRealm/}{purplehihi.github.io/TrajRealm/}}%
}
\date{}

\begin{document}
\maketitle
\input{sec/0_abstract}

\par\medskip
\noindent\begin{minipage}{\textwidth}
  \centering
  \includegraphics[width=\textwidth]{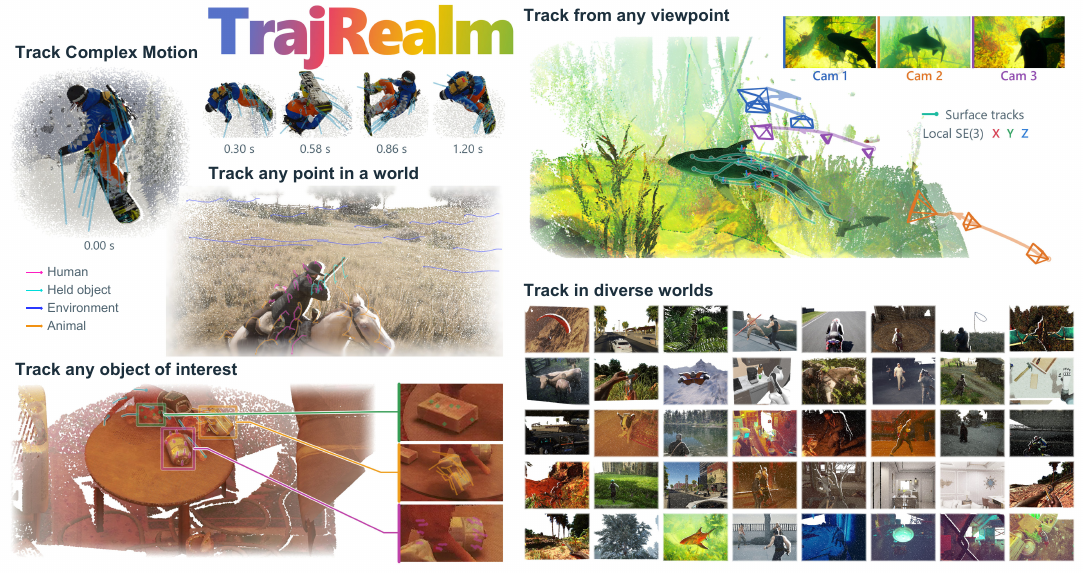}
  \captionsetup{hypcap=false}
  \captionof{figure}{\textbf{Highlights of TrajRealm.}
      TrajRealm covers complex motion, diverse targets and worlds, and multiple
      viewpoints, with point trajectories on foreground objects and environmental
      surfaces.}
  \label{fig:trajrealm-showcase}
\end{minipage}\par
\smallskip

\input{sec/1_intro}
\input{sec/2_related}
\input{sec/3_pipeline}

\input{sec/4_analysis}
\input{sec/5_experiments}
\input{sec/6_conclusion}
\FloatBarrier
\bibliographystyle{ieeenat_fullname_citeorder}
\bibliography{main}
\end{document}

%% file: preamble.tex
\usepackage{amsmath}
\usepackage{amssymb}
\usepackage{amsfonts}
\usepackage{graphicx}
\usepackage{booktabs}
\usepackage{multirow}
\usepackage{makecell}
\usepackage{subcaption}
\usepackage{nicefrac}
\usepackage{xcolor}
\usepackage{colortbl}
\usepackage{array}          
\usepackage{tabularx}       
\usepackage{placeins}   

\definecolor{LightGreen}{RGB}{59,125,35}

\definecolor{lightcyan}{rgb}{0.88, 1.0, 1.0}

\definecolor{myclay}{RGB}{59,182,176}
\definecolor{mypink}{RGB}{254,102,140}

\usepackage{xspace}
\newcommand{\paper}{TrajRealm\xspace}

\ifdefined\pdfcompresslevel
\fi

\makeatletter
\renewcommand\paragraph{\@startsection{paragraph}{4}{\z@}%
  {3pt}{-1em}{\normalfont\normalsize\bfseries}}
\makeatother
\AtBeginDocument{\raggedbottom}

%% file: sec/0_abstract.tex
\begin{abstract}
General-purpose point tracking requires large-scale training data with diverse dynamics, yet existing datasets struggle to combine rich dynamic content, accurate scene geometry, and sufficient scale. We introduce \textbf{TrajRealm}, the first public synthetic point-tracking dataset to reach the ten-million-frame scale, constructed from games and embodied simulations. TrajRealm uses the rich environments and motion systems of modern games to capture diverse targets, activities, and interactions across diverse worlds, while embodied simulations add robot manipulation and indoor navigation data. The dataset provides RGB observations, depth, camera parameters, and persistent 2D and 3D point trajectories. These annotations provide precise, geometrically consistent supervision across foreground objects and static environmental surfaces. We also build \textbf{TrajRealm-Bench} with eight subcategories under Interaction and Locomotion to evaluate tracking across diverse scenes. Benchmark evaluations reveal the strengths and limitations of existing trackers in handling complex motion and maintaining long-term point correspondences. Training on TrajRealm improves the performance of existing tracking methods on synthetic and real-world benchmarks.
\end{abstract}

%% file: sec/1_intro.tex
\section{Introduction}
\label{sec:introduction}

Point tracking follows the same physical points across video frames, providing correspondences for understanding object motion, scene dynamics, and 3D structure~\cite{harley2022pips,doersch2022tapvid}. Despite recent advances in general-purpose trackers~\cite{karaev2025cotracker3,zhang2025tapip3d}, interactions, deformation, occlusion, and camera motion still make accurate tracking difficult. Further progress requires large-scale training data covering diverse motions, interactions, and environments.

To address this need, early work explored manual annotation of point trajectories in real-world videos~\cite{doersch2022tapvid}. Although these videos capture diverse dynamics and scenes, annotating trajectories and obtaining depth and camera poses for 3D supervision are costly, making such datasets difficult to scale~\cite{koppula2024tapvid3d}. Building dynamic scenes in rendering engines enables large-scale data generation with ground-truth 3D point trajectories~\cite{zhao2026synthverse,jiang2026syn4d,greff2022kubric,zheng2023pointodyssey,jiang2026omnix}. However, reproducing diverse real-world scenes and motions remains difficult. The resulting data may contain unnatural motion, assets with limited detail, and outdoor backgrounds represented by simple textured surfaces rather than full 3D geometry~\cite{zhao2026synthverse,greff2022kubric,zheng2023pointodyssey}.

Our key observation is that modern games can offer more detailed 3D assets, advanced rendering, and richer motions and interactions than typical simulator scenes, making them a promising source of point-tracking data~\cite{richter2016playing,zhou2025omniworld}. Many feature connected worlds with detailed environments, realistic lighting, shadows, and changing weather. Each game offers multiple dynamic scenes, while different games broaden the range of targets, motions, and environments. However, robot manipulation and indoor navigation are less common in games, making embodied simulation data another valuable source for a comprehensive point-tracking dataset~\cite{chen2025robotwin,sun2026omninavbench}.

Based on these observations, we introduce \textbf{TrajRealm}, a ten-million-frame point-tracking dataset built from game and embodied simulation data. It pairs RGB observations with depth maps, camera parameters, and persistent 2D and 3D point trajectories. As shown in Fig.~\ref{fig:trajrealm-showcase}, TrajRealm covers complex motion, diverse targets and virtual worlds, and different viewpoints. Its geometrically consistent trajectories align with RGB observations and cover foreground objects and environmental surfaces. It includes single-view and multiview sequences and provides local SE(3) annotations~\cite{cheng2026mose3}.

To evaluate trackers across different dynamic scenes, we build the \textbf{TrajRealm-Bench} with 104,306  frames.
We organize its scenes into \textbf{Interaction} and \textbf{Locomotion} based on their main motion patterns.
Interaction focuses on relative motion among subjects, objects, and their parts.
Contact, joint motion, and occlusion make it difficult to track the same physical points over time.
Locomotion focuses on subjects or observers moving through environments and the resulting changes in visible surroundings.
Each category has four subcategories based on tracking challenges, targets, and scene settings.
Comparisons across these subcategories reveal differences that overall averages can hide.

Benchmark evaluations reveal complementary strengths among trackers. Trackers perform best on Embodied, while Hard and Long-horizon yield the lowest APD$_{2D}$ scores. CoTracker3~\cite{karaev2025cotracker3} outperforms TAPIP3D~\cite{zhang2025tapip3d} on Hard and Long-horizon, while TAPIP3D achieves higher 3D accuracy on Group and Plant, highlighting the importance of 3D geometric information. Fine-tuning TAPIP3D on TrajRealm also improves performance on synthetic and real-world benchmarks, with camera-space AJ$_{3D}$ gains of up to 7.83 percentage points relative to its published baseline. TrackEverything also improves all the average metrics on three real-world minival subsets, demonstrating the value of TrajRealm for training different tracking methods.

Our contributions are summarized as follows:
\begin{itemize}
    \item \textbf{A large-scale dataset with diverse dynamics.}
    TrajRealm combines game and embodied simulation data at the ten-million-frame scale, covering diverse motions and interactions with depth and point trajectories for both foreground objects and environmental surfaces.

    \item \textbf{A diverse tracking benchmark.}
    We establish a benchmark with eight subcategories under Interaction and Locomotion to evaluate tracking across different motions, targets, and environments.

    \item \textbf{Systematic experiments and analysis.}
    We analyze the strengths and limitations of existing trackers and show that training on TrajRealm improves performance on synthetic and real-world benchmarks.
\end{itemize}

%% file: sec/2_related.tex
\section{Related Work}
\label{sec:related}

\begin{table}[t]
\centering
\caption{\textbf{Point-tracking datasets and benchmarks.} Ego./Allo.: egocentric/allocentric.}
\label{tab:dataset_comparison}

\footnotesize
\setlength{\tabcolsep}{1.6pt}
\renewcommand{\arraystretch}{1.12}

\begin{tabular*}{\textwidth}{@{}l@{\hspace{5pt}}l@{\extracolsep{\fill}}cccccc@{}}
\specialrule{\heavyrulewidth}{0pt}{1.5pt}
Dataset & Source & Train Frames & Test Frames & Total Frames &
Content & Pipeline & View \\
\specialrule{\lightrulewidth}{1pt}{\belowrulesep}

Panoptic Studio~\cite{joo2015panoptic} & Real & $\times$ & 15K & 15K & Human & -- & -- \\
ADT~\cite{pan2023adt} & Real & $\times$ & 571K & 571K & Indoor & -- & Ego. \\
DriveTrack~\cite{balasingam2024drivetrack} & Real & -- & -- & 756K & Driving & -- & Ego. \\
TAPVid-3D~\cite{koppula2024tapvid3d} & Real & $\times$ & 600K & 600K & Diverse & -- & Ego. / Allo. \\

TAPVid-MV~\cite{tapvidmv2026} & Mixed & -- & -- & 207K & Diverse & Mixed &
Ego. / Allo. \\

Kubric~\cite{greff2022kubric} & Synthetic & 60K & 2.4K & 62K &
Falling objects & Blender & Allo. \\

PointOdyssey~\cite{zheng2023pointodyssey} & Synthetic & 166K & 26K & 216K &
Humans Animals & Blender & Ego. / Allo. \\

Dynamic Replica~\cite{karaev2023dynamicreplica} & Synthetic & 145K & 18K & 169.2K &
Humans Animals & Blender & Allo. \\

SynthVerse~\cite{zhao2026synthverse} & Synthetic & 5,816K & 59K & 5,875K &
Diverse &
Blender / Isaac Sim &
Ego. / Allo. \\

Art-Kubric~\cite{cheng2026mose3} & Synthetic & -- & -- & 600K &
Rigid / articulated & SAPIEN & Allo. \\

Syn4D~\cite{jiang2026syn4d} & Synthetic & -- & -- & 1,400K &
Diverse & Unreal Engine 5 & Allo. \\

\textbf{TrajRealm (Ours)} & Synthetic &
\textbf{9,896K} &
\textbf{104K} &
\textbf{10,000K} &
Diverse &
Games / Multiple Simulators &
Ego. / Allo. \\

\specialrule{\heavyrulewidth}{1pt}{0pt}
\end{tabular*}
\end{table}

\subsection{Point Tracking Methods}
\label{sec:related-methods}

Learning-based point tracking estimates persistent correspondences through occlusion and complex motion. PIPS~\cite{harley2022pips} and TAPIR~\cite{doersch2023tapir} estimate trajectories for queried points, while CoTracker3 combines joint tracking with training on pseudo-labeled real videos~\cite{karaev2025cotracker3}. Geometry extends this task beyond image coordinates: DELTA~\cite{ngo2025delta} estimates image positions and depth, SpatialTrackerV2~\cite{xiao2025spatialtrackerv2} models camera motion for 3D tracking, and TAPIP3D~\cite{zhang2025tapip3d} constructs correspondence features in persistent 3D space.

Recent work connects tracking with scene geometry and motion structure. D4RT uses a shared video representation and independent queries to recover 3D positions across time and camera coordinate systems~\cite{zhang2026d4rt}. OmniX separates motion from static geometry and predicts dense trajectory fields across views and time from compact dynamic tokens~\cite{jiang2026omnix}. MoSE3 jointly learns 3D tracks and rigidity embeddings, then differentiably fits world-space rigid transformations within soft rigid clusters~\cite{cheng2026mose3}.

Long videos and flexible observation sets introduce further challenges. TrackEverything maintains persistent world-space scene tracks and merges redundant tracks across temporal windows for dense tracking over long videos~\cite{jain2026trackeverything}. ARROW uses order-invariant queries to jointly reconstruct and track points from observations captured at different viewpoints and times~\cite{fradlin2026arrow}.

\subsection{Point-Tracking Datasets}
\label{sec:related-datasets}

Table~\ref{tab:dataset_comparison} summarizes representative point-tracking datasets and benchmarks. TAP-Vid~\cite{doersch2022tapvid} and TAPVid-3D~\cite{koppula2024tapvid3d} provide real-world benchmarks for 2D and 3D tracking, respectively. AnthroTAP~\cite{kim2026anthrotap} expands real-video training supervision by fitting parametric human models and filtering the resulting pseudo-labeled trajectories. Synthetic datasets provide direct access to scene state: Kubric~\cite{greff2022kubric} supports controlled scene generation, while PointOdyssey~\cite{zheng2023pointodyssey} emphasizes long trajectories with naturalistic motion.

Recent datasets broaden scene and motion coverage. SynthVerse~\cite{zhao2026synthverse} combines animated-film, embodied manipulation, navigation, and articulated objects. It is currently the largest available synthetic tracking dataset, but most of its data lacks full background geometry. MV-TAP~\cite{koo2026mvtap} introduces synthetic training data and real evaluation sets for cross-view tracking, while Art-Kubric~\cite{cheng2026mose3} adds articulated interactions with part-level motion and rigidity labels.

\begin{figure}[!t]
\centering
\includegraphics[width=\textwidth]{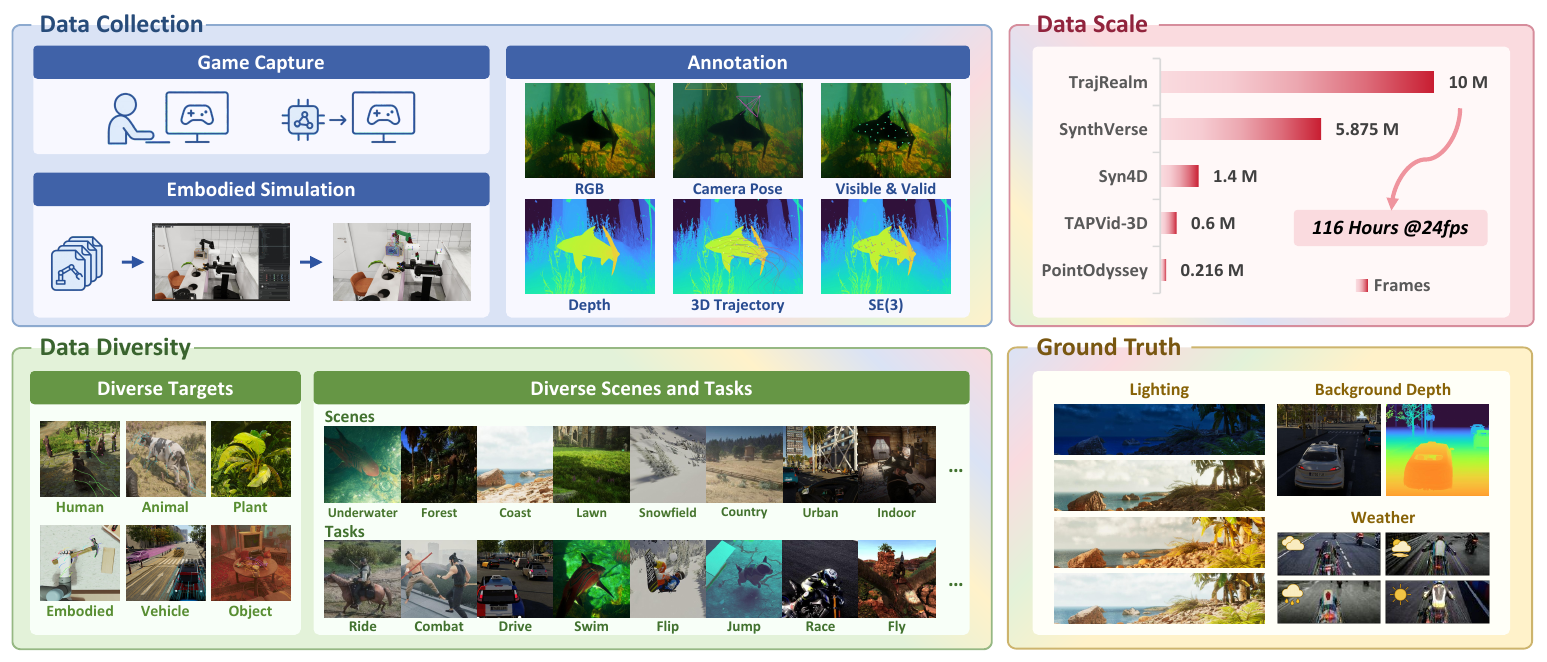}
\caption{\textbf{Data overview.}
Game capture and embodied re-rendering produce unified tracking data.
The panels summarize data collection, annotations, scale, diversity,
and visual and geometric quality.}
\label{fig:trajrealm-overview}
\end{figure}

OmniX~\cite{jiang2026omnix} and Syn4D~\cite{jiang2026syn4d} use Unreal Engine to combine animated assets with 3D environments, providing background depth and trajectories for static scene points. OmniX releases its data engine and example assets, but its full training and evaluation datasets and complete scene asset collection remain unreleased. Syn4D releases 1.4M frames across 30 3D environments. TrajRealm draws on realistic game worlds with integrated environments and activities, supplemented by embodied simulations. It provides tracking data at the ten-million-frame scale, with more diverse scenes and motions and consistent geometry across foreground objects and static environmental surfaces.

%% file: sec/3_pipeline.tex
\section{Data Collection and Analysis}

Figure~\ref{fig:trajrealm-overview} presents an overview of TrajRealm. The dataset combines two complementary sources, game sequences (TrajRealm-Game) and embodied simulations (TrajRealm-Embodied), in a unified representation of visual observations, persistent point trajectories, and geometric annotations.

\subsection{Data Source and Collection}

\paragraph{Game Tracking Data.}
We collect sequences directly from running games through manual play or automated control. For games with simple controls, we use built-in game AI or external navigation scripts. ReShade-based plugins capture rendering outputs, and plugins tailored to each game engine collect point trajectories without extracting the games' original 3D asset files. For selected games with unstable visual quality, we also use Qwen3.5-Flash to help screen the captured sequences.

\paragraph{Embodied Simulation Data.}
The embodied component draws on selected public datasets for robotic manipulation and indoor navigation. We re-render sequences from these datasets and convert the outputs into the same data representation as the game sequences. Table~\ref{tab:embodied} summarizes the selected datasets, simulation platforms, data types, and sequence and frame counts. These statistics describe only the selected and re-rendered subsets included in TrajRealm, rather than the full source datasets. For multiview data, synchronized frames count each scene state once, while camera frames count observations from all recorded views.

To prioritize task coverage while limiting redundancy, we include all available task types in each dataset but re-render at most 20 sequences per task across its perturbation variants. For datasets assembled from multiple sources, each source subset is screened separately. For example, we exclude the lower-quality Matterport-derived subset of OmniNavBench~\cite{sun2026omninavbench} and retain its GRScenes data.

\begin{table}[!t]
\centering
\caption{\textbf{Embodied data used in TrajRealm.}}
\label{tab:embodied}
\scriptsize
\setlength{\tabcolsep}{1pt}
\renewcommand{\arraystretch}{1.12}
\begin{tabular*}{\columnwidth}{@{\extracolsep{\fill}}>{\raggedright\arraybackslash}p{0.29\columnwidth}>{\centering\arraybackslash}p{0.14\columnwidth}>{\raggedleft\arraybackslash}p{0.09\columnwidth}>{\raggedleft\arraybackslash}p{0.22\columnwidth}>{\raggedleft\arraybackslash}p{0.22\columnwidth}@{}}
\specialrule{\heavyrulewidth}{0pt}{1.5pt}
\parbox[c]{\linewidth}{\raggedright Data Source} &
\parbox[c]{\linewidth}{\centering Simulator} &
\parbox[c]{\linewidth}{\raggedleft \mbox{Seq.}} &
\parbox[c]{\linewidth}{\raggedleft \mbox{Synch. frames}} &
\parbox[c]{\linewidth}{\raggedleft Camera frames} \\
\specialrule{\lightrulewidth}{1pt}{\belowrulesep}
\multicolumn{5}{l}{\textit{Gripper}} \\
EBench~\cite{gao2026ebench} & Isaac Sim & 76 & 165,845 & 331,690 \\
LIBERO-Plus~\cite{fei25libero-plus,liu2023libero} & MuJoCo & 120 & 24,639 & 49,278 \\
RoboTwin 2.0~\cite{chen2025robotwin} & SAPIEN & 935 & 271,836 & 271,836 \\
RoboCasa365~\cite{robocasa365} & MuJoCo & 607 & 478,216 & 1,434,648 \\
RoboDojo~\cite{chen2026robodojo} & Isaac Sim & 467 & 92,846 & 278,538 \\
\addlinespace[2pt]
\multicolumn{5}{l}{\textit{Dexterous Hand}} \\
Bench2Dex~\cite{yang2026bench2dex} & Isaac Sim & 78 & 57,728 & 230,912 \\
DexMimicGen~\cite{jiang2025dexmimicgen} & MuJoCo & 18 & 5,612 & 16,836 \\
\addlinespace[2pt]
\multicolumn{5}{l}{\textit{Navigation}} \\
InteriorNav~\cite{InteriorAgent2025} & Isaac Sim & 63 & 26,044 & 26,044 \\
OmniNavBench~\cite{sun2026omninavbench} & Isaac Sim & 48 & 254,862 & 254,862 \\
\specialrule{\lightrulewidth}{\aboverulesep}{1.5pt}
\textbf{Total} &  & \textbf{2,412} & \textbf{1,377,628} & \textbf{2,894,644} \\
\specialrule{\heavyrulewidth}{1pt}{0pt}
\end{tabular*}
\end{table}


\subsection{Data Annotation}

TrajRealm provides geometric ground truth that links RGB observations, 3D scene geometry, and persistent point trajectories. Within each sequence, the annotations share a consistent spatial and temporal reference, so that 3D positions, image projections, and available visibility labels describe the same scene state.

\paragraph{Observations, depth, and cameras.}
RGB frames and their aligned depth maps are indexed by sequence timestamps and camera identifiers. Camera intrinsics define the projection from 3D camera coordinates to the image plane, while extrinsics specify the transformations between camera and world coordinates.

\paragraph{Point trajectories and visibility.}
Persistent point identifiers link the same physical point across frames, including during motion and deformation. The trajectories cover both foreground objects and environmental surfaces (Fig.~\ref{fig:background-trajectories}(a)), with each 3D trajectory paired with its 2D projections in the available views. Separate masks record coordinate validity and visibility for each point and view, while an availability mask indicates whether a visibility label is provided. Reprojection, depth agreement, and cross-view consistency are used to check the geometric alignment between point trajectories and RGB observations.

\paragraph{Local SE(3) motion.}
For a subset of the data, we additionally provide local SE(3) motion annotations following the representation introduced by MoSE3~\cite{cheng2026mose3}. For a query frame $q$ and a target frame $t$, the transformation $T_{q\rightarrow t}=(R_{q\rightarrow t},t_{q\rightarrow t})\in\mathrm{SE}(3)$ describes the local motion associated with the query point in world coordinates. The rotation describes the motion of the associated rigid part or local surface, rather than an orientation of the point itself. Points on the same rigid part share a transform, while deforming surfaces are approximated by local rigid transforms. A validity mask identifies the supported point--frame pairs. These annotations complement point trajectories with local rotational information.

%% file: sec/4_analysis.tex
\begin{figure}[!t]
\centering
\includegraphics[width=\textwidth]{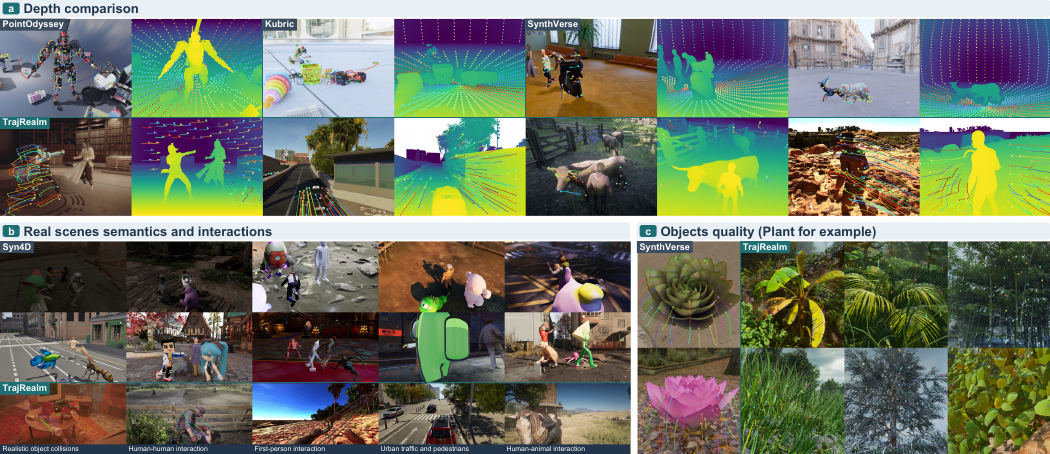}
\caption{\textbf{Comparison with synthetic datasets.}
(a) RGB--depth pairs from PointOdyssey~\cite{zheng2023pointodyssey}, Kubric~\cite{greff2022kubric}, and SynthVerse~\cite{zhao2026synthverse} (top), and TrajRealm (bottom). RGB shows world-space foreground trajectories; depth shows camera-relative background points or trajectories.
(b) Scene semantics and interactions in Syn4D~\cite{jiang2026syn4d} and TrajRealm.
(c) Plant appearance and trajectories in SynthVerse and TrajRealm.
Depth colors are normalized per example; white indicates unavailable depth.}
\label{fig:dataset-quality}
\label{fig:plant-comparison}
\label{fig:background-trajectories}
\end{figure}

\begin{figure}[!t]
\centering
\includegraphics[width=\textwidth]{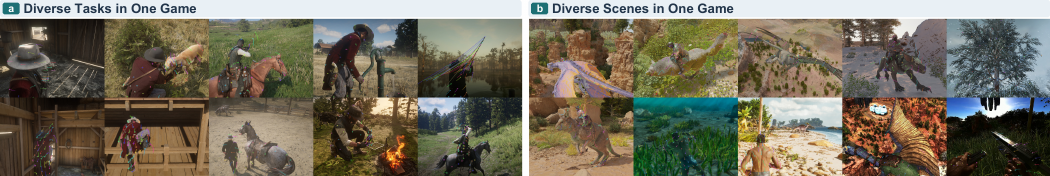}
\caption{\textbf{Within-game diversity.}
(a) Different activities within one game. (b) Diverse environments within another game.}
\label{fig:scene-diversity}
\end{figure}

\subsection{Data Analysis}

\label{sec:analysis}
\label{sec:analysis-diversity}

TrajRealm combines large-scale point-tracking data with diverse targets, environments, and dynamic activities, supported by realistic rendering, detailed depth, and RGB-aligned point trajectories.

\paragraph{Large Scale.}
As shown in Fig.~\ref{fig:trajrealm-overview}, TrajRealm comprises 13,972 sequences with 6.93M synchronized frames and 10M camera frames. TrajRealm-Game contributes 7.11M camera frames, while TrajRealm-Embodied contributes 2.89M. Local $\mathrm{SE}(3)$ motion annotations are available for 2,927 sequences covering 4,270,937 camera frames.

\paragraph{Target Diversity.}
TrajRealm covers humans, animals, plants, vehicles, robotic agents, and other objects (Fig.~\ref{fig:trajrealm-overview}). These targets include rigid objects, articulated bodies, and deformable structures. Figure~\ref{fig:plant-comparison}(c) compares plant appearances and trajectories in SynthVerse~\cite{zhao2026synthverse} and TrajRealm. The TrajRealm examples show finer leaf and branch structures and more natural shapes and textures, with tracking points distributed across the vegetation.

\paragraph{Diverse Dynamics.}
TrajRealm captures object manipulation, human--human and human--animal interactions, and movement through complex environments. Figure~\ref{fig:dataset-quality}(b) compares scene semantics and interactions in Syn4D~\cite{jiang2026syn4d} and TrajRealm, showing many tracking targets in Syn4D that do not fit their surroundings. Syn4D constructs diverse scenes by combining animated assets, but provides very limited coverage of semantically meaningful interactions and locomotion, with many objects performing simple motions. Its camera trajectories are also predominantly orbital. TrajRealm, in contrast, captures scene-grounded interactions and long-distance movement with cameras that follow the targets, providing sustained camera and object motion for training and evaluating trackers under moving-camera conditions. In TrajRealm, subjects and their actions fit the surrounding scene, spanning object collisions, combat, first-person manipulation, and urban traffic. These activities combine contact, rapid motion, pose changes, and occlusion, creating varied tracking challenges.

A single game provides a rich world of activities and environments. Figure~\ref{fig:scene-diversity} shows varied actions in one game (a) and forests, mountains, snowfields, and underwater environments in another (b). Combining multiple game worlds with embodied manipulation and navigation further broadens TrajRealm's scene and motion diversity.

\paragraph{Visual and Geometric Quality.}
TrajRealm combines realistic lighting, shadows, and weather (Fig.~\ref{fig:trajrealm-overview}) with detailed scene geometry. Figure~\ref{fig:background-trajectories}(a) compares RGB observations, depth, and trajectories from PointOdyssey~\cite{zheng2023pointodyssey}, Kubric~\cite{greff2022kubric}, SynthVerse~\cite{zhao2026synthverse}, and TrajRealm. The baseline examples have limited depth detail in their distant backgrounds. TrajRealm provides detailed depth and RGB-aligned trajectories on both foreground objects and environmental surfaces. The RGB panels show foreground trajectories in world coordinates, while the depth panels show background trajectories relative to the camera. Together, they capture object motion and the apparent motion of static surroundings as the camera moves.

\FloatBarrier

%% file: sec/5_experiments.tex
\section{Experiments}
\label{sec:experiments}

\subsection{Experimental Setup}
\label{sec:exp-setup}
\label{sec:exp-protocol}
\label{sec:benchmark-task}
\label{sec:benchmark-protocol}

\paragraph{Implementation details.}
Data rendering is performed on 64 NVIDIA RTX 4090 GPUs (48\,GB each). TAPIP3D and TrackEverything are fine-tuned for 30K steps using 16 NVIDIA A800 GPUs (80\,GB each).

\paragraph{Compared methods.}
Methods are grouped by GT depth and camera pose usage throughout the 3D tracking pipeline, including lifting and coordinate conversion. MoSE3, Track4World, and TrackEverything+$\pi^3$ use no GT geometry; CoTracker2.1, CoTracker3, DELTA, SpatialTrackerV2, TAPIP3D, and TrackEverything+GT use it. CoTracker tracks in RGB and uses GT depth and camera pose for 3D lifting; TrackEverything uses either GT geometry or $\pi^3$ estimates.

\paragraph{Evaluation metrics.}
Following SynthVerse~\cite{zhao2026synthverse} and TAPIP3D~\cite{zhang2025tapip3d}, we report AJ$_{3D}$, APD$_{3D}$, AJ$_{2D}$, APD$_{2D}$, and occlusion accuracy (OA). AJ measures joint localization and visibility accuracy, APD averages localization accuracy over five distance thresholds, and OA measures visibility classification accuracy. On TrajRealm-Bench, With GT methods use median scale normalization and depth-adaptive 3D thresholds. Without GT methods follow the MoSE3 protocol: first-frame camera coordinates, global L1 scale and translation fitting, a second median rescaling, and fixed 3D thresholds. Neither group fits rotation. All metrics favor higher scores; real-world benchmarks report AJ$_{3D}$, APD$_{3D}$, and OA.

\subsection{TrajRealm-Bench}
\label{sec:benchmark}
\label{sec:exp-trajrealm}

\paragraph{Benchmark Design.}
Point tracking in dynamic scenes involves different
correspondence challenges depending on how scene content
evolves over time. We organize dynamic scenes into two
fundamental categories according to their dominant motion
characteristics: Interaction and Locomotion. Interaction
focuses on relative motion among interacting subjects,
objects, and their constituent parts, where maintaining
point identity is essential despite contact, articulation,
and occlusion. Locomotion emphasizes the spatial movement
of subjects or observers through an environment and the
resulting changes in surrounding visible surfaces.
Within each category, we establish four subcategories.
General represents the basic setting of each category,
while the remaining subcategories extend the evaluation
to cover varying tracking difficulties, specialized
scenarios, and distinctive target properties. Table~\ref{tab:benchmark-splits}
summarizes the statistics of the eight evaluation subsets. 

\paragraph{Benchmark Categories.}
For Interaction, General covers everyday interactions
among humans, animals, and objects, evaluating point
correspondence as interacting subjects move relative to
one another. Hard extends this setting to high-dynamic
activities such as combat and skiing, involving rapid
motion, substantial pose changes, and frequent occlusion.
Embodied focuses on robotic manipulation, tracking points
on robotic grippers, dexterous hands, and manipulated
objects under contact and occlusion. Plant targets vegetation, where wind-driven motion
produces fine-scale deformation and changing
correspondences across leaves and branches. Compared
with existing plant tracking datasets, TrajRealm
provides more realistic plant appearances and motion,
captured in high-fidelity game environments.

For Locomotion, General covers individual movement by
humans and animals across terrestrial, aquatic, and aerial
environments, as well as embodied navigation where
surrounding environmental surfaces are the primary
tracking targets. Long-horizon extends this setting to
sequences with more than 800 temporal frames and sustained
movement, selected from general locomotion, transportation,
and group activities to evaluate long-term correspondence
preservation. During evaluation, some trackers exceeded GPU or host memory limits on Long-horizon. The full sequences provide a benchmark for future methods designed for extended tracking. To reduce evaluation cost, users can evaluate on Long-horizon Light, using only the first 800 frames of each sequence.
Transportation covers racetrack scenarios,
urban driving, and animal riding, focusing on the relative
motion of subjects, vehicles, and surrounding environments.
Group targets multiple moving humans or animals, evaluating
point correspondence as their trajectories intersect and
mutual occlusion occurs.

\begin{table}[htbp] 
    \centering
    \caption{\textbf{TrajRealm-Bench statistics.}}
    \label{tab:benchmark-splits}
    \footnotesize 
    \setlength{\tabcolsep}{0pt} 
    \renewcommand{\arraystretch}{0.85} 
    
    \begin{tabular}{@{} 
        >{\raggedright\arraybackslash}p{2.0cm}@{\hspace{4pt}} 
        >{\centering\arraybackslash}p{0.7cm}@{\hspace{4pt}}   
        >{\centering\arraybackslash}p{1.1cm}@{\hspace{1.5em}} 
        >{\raggedright\arraybackslash}p{2.0cm}@{\hspace{4pt}} 
        >{\centering\arraybackslash}p{0.7cm}@{\hspace{4pt}}   
        >{\centering\arraybackslash}p{1.1cm}                  
    @{}}
        \toprule
        \multicolumn{3}{c}{\textbf{Interaction}} & \multicolumn{3}{c}{\textbf{Locomotion}} \\
        \midrule 
        Category & Seq. & Frames & Category & Seq. & Frames \\
        \midrule
        General  & 40 & 8,495 & General        & 30 & 7,066 \\
        Hard     & 15 & 3,904 & Long-horizon   & 20 & 61,718 \\
        Embodied & 15 & 6,123 & Transportation & 30 & 11,492 \\
        Plant    & 15 & 3,690 & Group          & 10 & 1,818 \\
        \midrule
        Subtotal & 85 & 22,212 & Subtotal & 90 & 82,094 \\ 
        \bottomrule
    \end{tabular}
    \par\vspace{-5pt} 
\end{table}

\begin{figure}[!t]
\centering
\includegraphics[width=\textwidth]{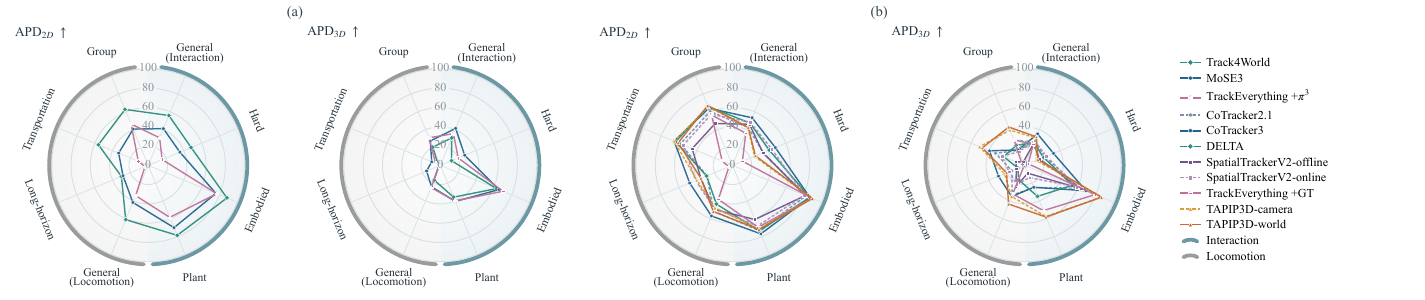}
\caption{\textbf{Performance across benchmark categories.} (a) Without GT depth and camera pose; (b) with GT depth and camera pose.}
\label{fig:benchmark-radar}
\end{figure}

\begin{table}[!t]
\centering
\caption{\textbf{TrajRealm-Bench results.} $^{*}$: fine-tuned; $^{\dagger}$: dense. GT groups include lifting and coordinate conversion; $\pi^3$~\cite{wang2026pi3}: estimated geometry. Track4World ($^{\ddagger}$) scores and averages exclude sequences with non-finite (NaN/Inf) predictions. Its successful/total sequence counts are Interaction---General 40/40, Hard 14/15, Embodied 14/15, Plant 15/15; Locomotion---General 29/30, Long-horizon 2/20, Transportation 14/30, Group 9/10.}
\label{tab:trajrealm-benchmark}
\setlength{\tabcolsep}{3pt}
\newcommand{\blankscore}{\makebox[2em]{}}
\renewcommand{\arraystretch}{1.05}
\newsavebox{\trajrealmbenchmarkpanel}
\sbox{\trajrealmbenchmarkpanel}{%
\begin{tabular}{l|ccccc|ccccc|ccccc|ccccc}
\toprule
\multirow{2}{*}{\shortstack[l]{\textbf{\textit{Interaction}}\\Methods}} & \multicolumn{5}{c|}{General} & \multicolumn{5}{c|}{Hard} & \multicolumn{5}{c|}{Embodied} & \multicolumn{5}{c}{Plant} \\
 & $\mathrm{AJ}_{3D}\uparrow$ & $\mathrm{APD}_{3D}\uparrow$ & $\mathrm{AJ}_{2D}\uparrow$ & $\mathrm{APD}_{2D}\uparrow$ & $\mathrm{OA}\uparrow$ & $\mathrm{AJ}_{3D}\uparrow$ & $\mathrm{APD}_{3D}\uparrow$ & $\mathrm{AJ}_{2D}\uparrow$ & $\mathrm{APD}_{2D}\uparrow$ & $\mathrm{OA}\uparrow$ & $\mathrm{AJ}_{3D}\uparrow$ & $\mathrm{APD}_{3D}\uparrow$ & $\mathrm{AJ}_{2D}\uparrow$ & $\mathrm{APD}_{2D}\uparrow$ & $\mathrm{OA}\uparrow$ & $\mathrm{AJ}_{3D}\uparrow$ & $\mathrm{APD}_{3D}\uparrow$ & $\mathrm{AJ}_{2D}\uparrow$ & $\mathrm{APD}_{2D}\uparrow$ & $\mathrm{OA}\uparrow$ \\
\midrule
\multicolumn{21}{l}{\textbf{\textit{Without GT depth and camera pose}}} \\
Track4World$^{\dagger}$~\cite{lu2026track4world} & 16.21 & 30.78 & 34.03 & 55.95 & 77.65 & 3.93\rlap{$^{\ddagger}$} & 12.56\rlap{$^{\ddagger}$} & 20.58\rlap{$^{\ddagger}$} & 48.24\rlap{$^{\ddagger}$} & 69.74\rlap{$^{\ddagger}$} & 52.65\rlap{$^{\ddagger}$} & 63.10\rlap{$^{\ddagger}$} & 77.44\rlap{$^{\ddagger}$} & 88.42\rlap{$^{\ddagger}$} & 91.34\rlap{$^{\ddagger}$} & 19.70 & 35.75 & 44.20 & 78.60 & 60.78 \\
MoSE3$^{\dagger}$~\cite{cheng2026mose3} & 22.77 & 41.48 & 22.80 & 41.16 & 76.18 & 10.80 & 27.34 & 13.08 & 35.60 & 65.17 & 55.19 & 65.81 & 63.09 & 75.99 & 93.01 & 22.54 & 40.53 & 38.75 & 69.98 & 61.83 \\
{TrackEverything +$\pi^3$}$^{\dagger}$~\cite{jain2026trackeverything} & 16.18 & 35.57 & 15.28 & 30.89 & 64.13 & 6.56 & 20.06 & 4.62 & 15.67 & 46.86 & 56.64 & 70.33 & 62.02 & 76.97 & 89.59 & 21.79 & 39.88 & 31.95 & 58.40 & 63.65 \\
\specialrule{\lightrulewidth}{0.5pt}{0.5pt}
TrackEverything +$\pi^3$$^{*\dagger}$ & 21.70 & 40.91 & 24.20 & 44.13 & 78.18 & 10.11 & 26.52 & 12.74 & 32.35 & 68.25 & 55.92 & 69.22 & 59.89 & 74.49 & 89.64 & 21.28 & 39.12 & 34.56 & 63.36 & 65.35 \\
\addlinespace[2pt]
\multicolumn{21}{l}{\textbf{\textit{With GT depth and camera pose}}} \\
CoTracker2.1~\cite{karaev2024cotracker} & 14.58 & 27.48 & 22.01 & 42.51 & 68.67 & 8.67 & 22.80 & 12.69 & 34.41 & 62.46 & 50.22 & 61.62 & 68.67 & 82.57 & 84.29 & 13.19 & 24.98 & 37.47 & 73.37 & 58.40 \\
CoTracker3~\cite{karaev2025cotracker3} & 19.60 & 35.29 & 28.95 & 53.36 & 75.91 & 12.79 & 32.44 & 16.25 & 48.14 & 72.84 & 57.33 & 68.88 & 74.27 & 86.90 & 88.34 & 13.67 & 24.66 & 43.50 & 76.92 & 59.60 \\
DELTA$^{\dagger}$~\cite{ngo2025delta} & 14.66 & 25.60 & 27.88 & 47.47 & 72.59 & 8.29 & 21.30 & 15.78 & 41.28 & 58.15 & 43.77 & 53.91 & 76.05 & 86.65 & 91.58 & 21.22 & 34.86 & 44.95 & 73.65 & 70.23 \\
SpatialTrackerV2-offline~\cite{xiao2025spatialtrackerv2} & 14.84 & 24.28 & 23.88 & 46.07 & 69.26 & 5.48 & 16.74 & 11.33 & 34.23 & 55.60 & 51.59 & 63.86 & 63.45 & 81.79 & 83.75 & 4.01 & 8.85 & 32.98 & 60.81 & 59.55 \\
SpatialTrackerV2-online~\cite{xiao2025spatialtrackerv2} & 11.65 & 27.51 & 20.10 & 48.40 & 51.97 & 6.53 & 24.45 & 11.88 & 44.12 & 37.87 & 42.33 & 58.62 & 60.05 & 80.85 & 82.07 & 6.87 & 13.84 & 36.47 & 68.09 & 59.05 \\
TrackEverything +GT$^{\dagger}$~\cite{jain2026trackeverything} & 13.15 & 23.79 & 19.12 & 36.11 & 68.41 & 1.57 & 4.07 & 3.81 & 11.22 & 51.56 & 67.38 & 79.22 & 76.36 & 89.11 & 92.96 & 30.39 & 50.89 & 41.60 & 69.50 & 70.80 \\
TAPIP3D-camera~\cite{zhang2025tapip3d} & 14.62 & 30.65 & 19.45 & 42.37 & 53.34 & 4.05 & 13.31 & 7.36 & 24.59 & 36.82 & 70.10 & 86.17 & 73.93 & 89.41 & 90.56 & 32.42 & 57.49 & 40.29 & 71.51 & 58.70 \\
TAPIP3D-world~\cite{zhang2025tapip3d} & 15.74 & 31.98 & 20.06 & 42.61 & 55.35 & 4.38 & 15.07 & 7.60 & 25.83 & 36.19 & 69.97 & 86.35 & 74.32 & 89.75 & 90.66 & 32.08 & 58.04 & 40.15 & 71.99 & 58.29 \\
\specialrule{\lightrulewidth}{0.5pt}{0.5pt}
TAPIP3D-camera$^{*}$ & 32.03 & 46.98 & 38.67 & 59.36 & 82.53 & 22.56 & 41.58 & 27.18 & 54.72 & 76.74 & 74.48 & 86.25 & 77.35 & 90.01 & 91.90 & 44.16 & 66.56 & 49.79 & 79.47 & 72.66 \\
TAPIP3D-world$^{*}$ & 31.95 & 44.99 & 37.35 & 55.09 & 80.91 & 21.36 & 41.71 & 26.65 & 55.35 & 76.98 & 75.03 & 86.58 & 78.15 & 90.41 & 92.41 & 43.77 & 64.86 & 49.71 & 78.90 & 73.21 \\
\end{tabular}}
\resizebox{\textwidth}{!}{%
\begin{minipage}{\wd\trajrealmbenchmarkpanel}
\setlength{\parindent}{0pt}
\usebox{\trajrealmbenchmarkpanel}
\par\vspace{1pt}
\begin{tabular*}{\linewidth}{l|@{\hspace{\tabcolsep}\extracolsep{\fill}}ccccc|ccccc|ccccc|ccccc}
\midrule
\multirow{2}{*}{\shortstack[l]{\textbf{\textit{Locomotion}}\\Methods}} & \multicolumn{5}{c|}{General} & \multicolumn{5}{c|}{Long-horizon} & \multicolumn{5}{c|}{Transportation} & \multicolumn{5}{c}{Group} \\
 & $\mathrm{AJ}_{3D}\uparrow$ & $\mathrm{APD}_{3D}\uparrow$ & $\mathrm{AJ}_{2D}\uparrow$ & $\mathrm{APD}_{2D}\uparrow$ & $\mathrm{OA}\uparrow$ & $\mathrm{AJ}_{3D}\uparrow$ & $\mathrm{APD}_{3D}\uparrow$ & $\mathrm{AJ}_{2D}\uparrow$ & $\mathrm{APD}_{2D}\uparrow$ & $\mathrm{OA}\uparrow$ & $\mathrm{AJ}_{3D}\uparrow$ & $\mathrm{APD}_{3D}\uparrow$ & $\mathrm{AJ}_{2D}\uparrow$ & $\mathrm{APD}_{2D}\uparrow$ & $\mathrm{OA}\uparrow$ & $\mathrm{AJ}_{3D}\uparrow$ & $\mathrm{APD}_{3D}\uparrow$ & $\mathrm{AJ}_{2D}\uparrow$ & $\mathrm{APD}_{2D}\uparrow$ & $\mathrm{OA}\uparrow$ \\
\midrule
\multicolumn{21}{l}{\textbf{\textit{Without GT depth and camera pose}}} \\
Track4World$^{\dagger}$~\cite{lu2026track4world} & 12.04\rlap{$^{\ddagger}$} & 16.54\rlap{$^{\ddagger}$} & 41.39\rlap{$^{\ddagger}$} & 60.96\rlap{$^{\ddagger}$} & 79.90\rlap{$^{\ddagger}$} & 0.52\rlap{$^{\ddagger}$} & 1.91\rlap{$^{\ddagger}$} & 14.06\rlap{$^{\ddagger}$} & 30.07\rlap{$^{\ddagger}$} & 74.94\rlap{$^{\ddagger}$} & 1.00\rlap{$^{\ddagger}$} & 2.52\rlap{$^{\ddagger}$} & 30.86\rlap{$^{\ddagger}$} & 55.89\rlap{$^{\ddagger}$} & 70.61\rlap{$^{\ddagger}$} & 11.52\rlap{$^{\ddagger}$} & 20.38\rlap{$^{\ddagger}$} & 40.91\rlap{$^{\ddagger}$} & 62.76\rlap{$^{\ddagger}$} & 76.56\rlap{$^{\ddagger}$} \\
MoSE3$^{\dagger}$~\cite{cheng2026mose3} & 17.42 & 24.02 & 26.60 & 41.62 & 79.59 & 6.80 & 15.40 & 10.59 & 28.25 & 65.55 & 4.53 & 9.65 & 15.93 & 33.18 & 66.85 & 15.29 & 27.30 & 23.95 & 40.66 & 74.40 \\
{TrackEverything +$\pi^3$}$^{\dagger}$~\cite{jain2026trackeverything} & 19.03 & 25.06 & 25.43 & 34.01 & 72.43 & 0.42 & 1.72 & 1.56 & 3.64 & 61.67 & 2.59 & 5.28 & 5.55 & 11.14 & 58.95 & 15.73 & 30.90 & 23.63 & 46.41 & 66.64 \\
\specialrule{\lightrulewidth}{0.5pt}{0.5pt}
TrackEverything +$\pi^3$$^{*\dagger}$ & 21.27 & 29.09 & 30.38 & 43.99 & 79.02 & 1.73 & 4.75 & 5.36 & 14.49 & 66.86 & 5.13 & 10.15 & 14.37 & 29.22 & 65.20 & 17.33 & 32.20 & 29.30 & 53.62 & 70.90 \\
\addlinespace[2pt]
\multicolumn{21}{l}{\textbf{\textit{With GT depth and camera pose}}} \\
CoTracker2.1~\cite{karaev2024cotracker} & 20.98 & 33.41 & 32.31 & 51.40 & 75.08 & 6.86 & 14.40 & 13.09 & 28.62 & 62.57 & 17.08 & 32.60 & 31.11 & 58.30 & 69.57 & 9.11 & 16.78 & 34.40 & 60.53 & 71.95 \\
CoTracker3~\cite{karaev2025cotracker3} & 22.26 & 35.09 & 35.68 & 56.86 & 78.02 & 12.46 & 29.30 & 19.75 & 48.03 & 69.47 & 20.95 & 39.37 & 35.76 & 65.44 & 74.39 & 9.96 & 17.28 & 38.36 & 64.05 & 75.33 \\
DELTA$^{\dagger}$~\cite{ngo2025delta} & 10.16 & 16.14 & 30.32 & 43.51 & 74.02 & 3.30 & 7.52 & 13.29 & 28.95 & 61.84 & 11.66 & 23.49 & 35.81 & 63.88 & 70.07 & 12.24 & 22.46 & 40.16 & 66.01 & 76.36 \\
SpatialTrackerV2-offline~\cite{xiao2025spatialtrackerv2} & 16.69 & 24.05 & 30.78 & 49.55 & 74.61 & 3.59 & 9.07 & 11.65 & 35.43 & 47.54 & 5.06 & 9.78 & 19.86 & 45.17 & 60.27 & 2.47 & 3.56 & 24.62 & 46.98 & 68.20 \\
SpatialTrackerV2-online~\cite{xiao2025spatialtrackerv2} & 11.17 & 21.04 & 24.84 & 48.72 & 60.21 & 5.43 & 17.52 & 12.55 & 41.28 & 38.55 & 9.91 & 25.03 & 22.16 & 55.03 & 46.58 & 5.60 & 15.22 & 27.99 & 57.32 & 61.02 \\
TrackEverything +GT$^{\dagger}$~\cite{jain2026trackeverything} & 28.01 & 32.06 & 32.47 & 38.17 & 72.98 & 0.20 & 0.44 & 1.74 & 3.99 & 62.48 & 1.42 & 2.83 & 5.93 & 11.60 & 59.28 & 17.74 & 28.60 & 31.15 & 56.44 & 70.40 \\
TAPIP3D-camera~\cite{zhang2025tapip3d} & 23.13 & 35.00 & 30.16 & 47.55 & 65.94 & 8.31 & 22.63 & 13.69 & 39.96 & 46.29 & 24.91 & 50.47 & 31.44 & 65.38 & 56.86 & 22.90 & 39.54 & 35.28 & 65.72 & 67.16 \\
TAPIP3D-world~\cite{zhang2025tapip3d} & 30.11 & 43.44 & 34.91 & 51.86 & 67.53 & 7.41 & 20.19 & 12.44 & 35.37 & 49.34 & 23.15 & 47.38 & 29.05 & 60.94 & 56.23 & 23.56 & 42.98 & 36.14 & 67.03 & 67.24 \\
\specialrule{\lightrulewidth}{0.5pt}{0.5pt}
TAPIP3D-camera$^{*}$ & 38.23 & 53.51 & 46.91 & 66.54 & 84.77 & 27.24 & 42.04 & 33.50 & 58.54 & 80.44 & 44.06 & 63.77 & 51.51 & 75.68 & 84.94 & 39.14 & 54.88 & 50.14 & 76.16 & 81.38 \\
TAPIP3D-world$^{*}$ & 42.62 & 56.86 & 49.09 & 66.57 & 84.50 & 20.29 & 30.66 & 26.25 & 45.47 & 76.84 & 38.14 & 55.37 & 46.12 & 67.95 & 82.45 & 43.13 & 63.33 & 50.65 & 76.66 & 81.89 \\
\end{tabular*}
\par\vspace{1pt}
\begin{tabular*}{\linewidth}{l|@{\hspace{\tabcolsep}\extracolsep{\fill}}ccccc|ccccc|ccccc}
\midrule
\multirow{2}{*}{\shortstack[l]{\textbf{\textit{Averages}}\\Methods}} & \multicolumn{5}{c|}{Interaction Avg.} & \multicolumn{5}{c|}{Locomotion Avg.} & \multicolumn{5}{c}{Overall Avg.} \\
 & $\mathrm{AJ}_{3D}\uparrow$ & $\mathrm{APD}_{3D}\uparrow$ & $\mathrm{AJ}_{2D}\uparrow$ & $\mathrm{APD}_{2D}\uparrow$ & $\mathrm{OA}\uparrow$ & $\mathrm{AJ}_{3D}\uparrow$ & $\mathrm{APD}_{3D}\uparrow$ & $\mathrm{AJ}_{2D}\uparrow$ & $\mathrm{APD}_{2D}\uparrow$ & $\mathrm{OA}\uparrow$ & $\mathrm{AJ}_{3D}\uparrow$ & $\mathrm{APD}_{3D}\uparrow$ & $\mathrm{AJ}_{2D}\uparrow$ & $\mathrm{APD}_{2D}\uparrow$ & $\mathrm{OA}\uparrow$ \\
\midrule
\multicolumn{16}{l}{\textbf{\textit{Without GT depth and camera pose}}} \\
Track4World$^{\dagger}$~\cite{lu2026track4world} & 23.12\rlap{$^{\ddagger}$} & 35.55\rlap{$^{\ddagger}$} & 44.06\rlap{$^{\ddagger}$} & 67.80\rlap{$^{\ddagger}$} & 74.88\rlap{$^{\ddagger}$} & 6.27\rlap{$^{\ddagger}$} & 10.34\rlap{$^{\ddagger}$} & 31.81\rlap{$^{\ddagger}$} & 52.42\rlap{$^{\ddagger}$} & 75.50\rlap{$^{\ddagger}$} & 14.70\rlap{$^{\ddagger}$} & 22.94\rlap{$^{\ddagger}$} & 37.93\rlap{$^{\ddagger}$} & 60.11\rlap{$^{\ddagger}$} & 75.19\rlap{$^{\ddagger}$} \\
MoSE3$^{\dagger}$~\cite{cheng2026mose3} & 27.82 & 43.79 & 34.43 & 55.68 & 74.05 & 11.01 & 19.09 & 19.27 & 35.93 & 71.60 & 19.42 & 31.44 & 26.85 & 45.81 & 72.82 \\
{TrackEverything +$\pi^3$}$^{\dagger}$~\cite{jain2026trackeverything} & 25.29 & 41.46 & 28.47 & 45.48 & 66.06 & 9.44 & 15.74 & 14.04 & 23.80 & 64.92 & 17.37 & 28.60 & 21.26 & 34.64 & 65.49 \\
\specialrule{\lightrulewidth}{0.5pt}{0.5pt}
TrackEverything +$\pi^3$$^{*\dagger}$ & 27.25 & 43.94 & 32.85 & 53.58 & 75.35 & 11.37 & 19.05 & 19.85 & 35.33 & 70.49 & 19.31 & 31.50 & 26.35 & 44.46 & 72.92 \\
\addlinespace[2pt]
\multicolumn{16}{l}{\textbf{\textit{With GT depth and camera pose}}} \\
CoTracker2.1~\cite{karaev2024cotracker} & 21.66 & 34.22 & 35.21 & 58.22 & 68.46 & 13.51 & 24.30 & 27.73 & 49.71 & 69.79 & 17.59 & 29.26 & 31.47 & 53.96 & 69.12 \\
CoTracker3~\cite{karaev2025cotracker3} & 25.85 & 40.32 & 40.74 & 66.33 & 74.17 & 16.41 & 30.26 & 32.39 & 58.59 & 74.30 & 21.13 & 35.29 & 36.56 & 62.46 & 74.24 \\
DELTA$^{\dagger}$~\cite{ngo2025delta} & 21.99 & 33.92 & 41.17 & 62.26 & 73.14 & 9.34 & 17.40 & 29.90 & 50.59 & 70.57 & 15.66 & 25.66 & 35.53 & 56.43 & 71.86 \\
SpatialTrackerV2-offline~\cite{xiao2025spatialtrackerv2} & 18.98 & 28.43 & 32.91 & 55.73 & 67.04 & 6.95 & 11.61 & 21.73 & 44.28 & 62.65 & 12.97 & 20.02 & 27.32 & 50.01 & 64.85 \\
SpatialTrackerV2-online~\cite{xiao2025spatialtrackerv2} & 16.85 & 31.10 & 32.12 & 60.37 & 57.74 & 8.03 & 19.70 & 21.88 & 50.59 & 51.59 & 12.44 & 25.40 & 27.00 & 55.48 & 54.67 \\
TrackEverything +GT$^{\dagger}$~\cite{jain2026trackeverything} & 28.12 & 39.49 & 35.22 & 51.49 & 70.93 & 11.84 & 15.98 & 17.82 & 27.55 & 66.29 & 19.98 & 27.74 & 26.52 & 39.52 & 68.61 \\
TAPIP3D-camera~\cite{zhang2025tapip3d} & 30.30 & 46.90 & 35.26 & 56.97 & 59.85 & 19.81 & 36.91 & 27.64 & 54.65 & 59.07 & 25.06 & 41.91 & 31.45 & 55.81 & 59.46 \\
TAPIP3D-world~\cite{zhang2025tapip3d} & 30.54 & 47.86 & 35.53 & 57.55 & 60.12 & 21.06 & 38.50 & 28.13 & 53.80 & 60.08 & 25.80 & 43.18 & 31.83 & 55.67 & 60.10 \\
\specialrule{\lightrulewidth}{0.5pt}{0.5pt}
TAPIP3D-camera$^{*}$ & 43.31 & 60.34 & 48.25 & 70.89 & 80.96 & 37.17 & 53.55 & 45.51 & 69.23 & 82.88 & 40.24 & 56.95 & 46.88 & 70.06 & 81.92 \\
TAPIP3D-world$^{*}$ & 43.03 & 59.53 & 47.97 & 69.94 & 80.88 & 36.04 & 51.56 & 43.03 & 64.16 & 81.42 & 39.54 & 55.55 & 45.50 & 67.05 & 81.15 \\
\bottomrule
\end{tabular*}
\end{minipage}}
\end{table}

\begin{table}[!t]
\centering
\caption{\textbf{Public synthetic benchmark results.} $^{*}$: fine-tuned on \paper. $\Delta$: fine-tuned minus published baseline (pp). Light blue marks the higher score.}
\label{tab:synthetic_tracking_comparison}
\setlength{\tabcolsep}{3pt}
\renewcommand{\arraystretch}{1.05}
\resizebox{\textwidth}{!}{%
\begin{tabular}{l|ccccc|ccccc|ccccc|ccccc}
\toprule
\multirow{2}{*}{Methods} & \multicolumn{5}{c|}{SynthVerse-mAverage} & \multicolumn{5}{c|}{Dynamic Replica} & \multicolumn{5}{c|}{LSFOdyssey} & \multicolumn{5}{c}{Avg.} \\
 & $\mathrm{AJ}_{3D}\uparrow$ & $\mathrm{APD}_{3D}\uparrow$ & $\mathrm{AJ}_{2D}\uparrow$ & $\mathrm{APD}_{2D}\uparrow$ & $\mathrm{OA}\uparrow$ & $\mathrm{AJ}_{3D}\uparrow$ & $\mathrm{APD}_{3D}\uparrow$ & $\mathrm{AJ}_{2D}\uparrow$ & $\mathrm{APD}_{2D}\uparrow$ & $\mathrm{OA}\uparrow$ & $\mathrm{AJ}_{3D}\uparrow$ & $\mathrm{APD}_{3D}\uparrow$ & $\mathrm{AJ}_{2D}\uparrow$ & $\mathrm{APD}_{2D}\uparrow$ & $\mathrm{OA}\uparrow$ & $\mathrm{AJ}_{3D}\uparrow$ & $\mathrm{APD}_{3D}\uparrow$ & $\mathrm{AJ}_{2D}\uparrow$ & $\mathrm{APD}_{2D}\uparrow$ & $\mathrm{OA}\uparrow$ \\

\midrule
\multicolumn{21}{l}{\textit{TAPIP3D-camera}} \\
TAPIP3D & 33.30 & 46.20 & 47.10 & \cellcolor{blue!8}66.90 & 79.40 & 53.70 & 70.80 & 64.60 & 84.70 & 84.70 & 68.30 & 83.20 & 76.00 & 91.20 & 86.20 & 51.80 & 66.70 & 62.60 & 80.90 & 83.40 \\
TAPIP3D$^{*}$ & \cellcolor{blue!8}36.41 & \cellcolor{blue!8}48.01 & \cellcolor{blue!8}48.10 & 65.91 & \cellcolor{blue!8}79.91 & \cellcolor{blue!8}61.53 & \cellcolor{blue!8}75.42 & \cellcolor{blue!8}69.76 & \cellcolor{blue!8}85.43 & \cellcolor{blue!8}87.81 & \cellcolor{blue!8}73.53 & \cellcolor{blue!8}86.04 & \cellcolor{blue!8}80.16 & \cellcolor{blue!8}92.35 & \cellcolor{blue!8}90.42 & \cellcolor{blue!8}57.16 & \cellcolor{blue!8}69.82 & \cellcolor{blue!8}66.01 & \cellcolor{blue!8}81.23 & \cellcolor{blue!8}86.05 \\
$\Delta$ & $+3.11$ & $+1.81$ & $+1.00$ & $-0.99$ & $+0.51$ & $+7.83$ & $+4.62$ & $+5.16$ & $+0.73$ & $+3.11$ & $+5.23$ & $+2.84$ & $+4.16$ & $+1.15$ & $+4.22$ & $+5.36$ & $+3.12$ & $+3.41$ & $+0.33$ & $+2.65$ \\
\midrule
\multicolumn{21}{l}{\textit{TAPIP3D-world}} \\
TAPIP3D & 33.30 & 46.40 & 47.40 & \cellcolor{blue!8}67.20 & 79.80 & 55.50 & 72.80 & 66.20 & 85.70 & 85.30 & 72.20 & 85.80 & 78.50 & 92.80 & 86.90 & 53.70 & 68.30 & 64.00 & 81.90 & 84.00 \\
TAPIP3D$^{*}$ & \cellcolor{blue!8}37.59 & \cellcolor{blue!8}49.41 & \cellcolor{blue!8}49.08 & 66.62 & \cellcolor{blue!8}80.81 & \cellcolor{blue!8}62.80 & \cellcolor{blue!8}75.71 & \cellcolor{blue!8}71.09 & \cellcolor{blue!8}86.59 & \cellcolor{blue!8}87.53 & \cellcolor{blue!8}76.24 & \cellcolor{blue!8}88.08 & \cellcolor{blue!8}82.18 & \cellcolor{blue!8}93.72 & \cellcolor{blue!8}90.54 & \cellcolor{blue!8}58.88 & \cellcolor{blue!8}71.07 & \cellcolor{blue!8}67.45 & \cellcolor{blue!8}82.31 & \cellcolor{blue!8}86.29 \\
$\Delta$ & $+4.29$ & $+3.01$ & $+1.68$ & $-0.58$ & $+1.01$ & $+7.30$ & $+2.91$ & $+4.89$ & $+0.89$ & $+2.23$ & $+4.04$ & $+2.28$ & $+3.68$ & $+0.92$ & $+3.64$ & $+5.18$ & $+2.77$ & $+3.45$ & $+0.41$ & $+2.29$ \\
\bottomrule
\end{tabular}}
\end{table}

\begin{table}[!t]
\centering
\caption{\textbf{Public real-world benchmark results.} $^{*}$: fine-tuned on \paper. TrackEverything~\cite{jain2026trackeverything} uses 50 full-length minival videos per dataset. $\Delta$: fine-tuned minus published baseline (pp). Light blue marks the higher score.}
\label{tab:tracking_comparison}
\setlength{\tabcolsep}{1.5pt}
\renewcommand{\arraystretch}{1.05}
\resizebox{\columnwidth}{!}{%
\begin{tabular}{l|ccc|ccc|ccc|ccc}
\toprule
\multirow{2}{*}{Methods} & \multicolumn{3}{c|}{ADT} & \multicolumn{3}{c|}{DriveTrack} & \multicolumn{3}{c|}{PStudio} & \multicolumn{3}{c}{Avg.} \\
 & $\mathrm{AJ}_{3D}\uparrow$ & $\mathrm{APD}_{3D}\uparrow$ & $\mathrm{OA}\uparrow$ & $\mathrm{AJ}_{3D}\uparrow$ & $\mathrm{APD}_{3D}\uparrow$ & $\mathrm{OA}\uparrow$ & $\mathrm{AJ}_{3D}\uparrow$ & $\mathrm{APD}_{3D}\uparrow$ & $\mathrm{OA}\uparrow$ & $\mathrm{AJ}_{3D}\uparrow$ & $\mathrm{APD}_{3D}\uparrow$ & $\mathrm{OA}\uparrow$ \\

\midrule
\multicolumn{13}{l}{\textit{TAPIP3D-camera +MegaSaM}} \\
TAPIP3D & 21.60 & 31.00 & 90.40 & 14.60 & \cellcolor{blue!8}21.30 & 82.20 & 18.10 & 27.70 & 85.50 & 18.10 & 26.70 & 86.00 \\
TAPIP3D$^{*}$ & \cellcolor{blue!8}24.00 & \cellcolor{blue!8}32.55 & \cellcolor{blue!8}93.55 & \cellcolor{blue!8}14.80 & 20.65 & \cellcolor{blue!8}84.34 & \cellcolor{blue!8}19.06 & \cellcolor{blue!8}28.52 & \cellcolor{blue!8}88.04 & \cellcolor{blue!8}19.29 & \cellcolor{blue!8}27.24 & \cellcolor{blue!8}88.64 \\
$\Delta$ & $+2.40$ & $+1.55$ & $+3.15$ & $+0.20$ & $-0.65$ & $+2.14$ & $+0.96$ & $+0.82$ & $+2.54$ & $+1.19$ & $+0.54$ & $+2.64$ \\
\midrule
\multicolumn{13}{l}{\textit{TAPIP3D-world +MegaSaM}} \\
TAPIP3D & 23.50 & 32.80 & 91.20 & 14.90 & 21.80 & 82.60 & 18.10 & 27.70 & 85.50 & 18.80 & 27.40 & 86.40 \\
TAPIP3D$^{*}$ & \cellcolor{blue!8}25.39 & \cellcolor{blue!8}34.14 & \cellcolor{blue!8}94.08 & \cellcolor{blue!8}15.51 & \cellcolor{blue!8}22.12 & \cellcolor{blue!8}84.17 & \cellcolor{blue!8}19.05 & \cellcolor{blue!8}28.49 & \cellcolor{blue!8}88.03 & \cellcolor{blue!8}19.98 & \cellcolor{blue!8}28.25 & \cellcolor{blue!8}88.76 \\
$\Delta$ & $+1.89$ & $+1.34$ & $+2.88$ & $+0.61$ & $+0.32$ & $+1.57$ & $+0.95$ & $+0.79$ & $+2.53$ & $+1.18$ & $+0.85$ & $+2.36$ \\
\midrule
\multicolumn{13}{l}{\textit{TrackEverything +$\pi^3$}} \\
TrackEverything & 21.10 & 31.00 & 87.90 & 17.30 & 24.70 & 84.20 & 11.40 & 18.90 & \cellcolor{blue!8}86.00 & 16.60 & 24.90 & 86.00 \\
TrackEverything$^{*}$ & \cellcolor{blue!8}24.93 & \cellcolor{blue!8}32.79 & \cellcolor{blue!8}91.03 & \cellcolor{blue!8}18.84 & \cellcolor{blue!8}24.92 & \cellcolor{blue!8}89.26 & \cellcolor{blue!8}12.89 & \cellcolor{blue!8}20.51 & 85.39 & \cellcolor{blue!8}18.88 & \cellcolor{blue!8}26.08 & \cellcolor{blue!8}88.56 \\
$\Delta$ & $+3.83$ & $+1.79$ & $+3.13$ & $+1.54$ & $+0.22$ & $+5.06$ & $+1.49$ & $+1.61$ & $-0.61$ & $+2.28$ & $+1.18$ & $+2.56$ \\
\bottomrule
\end{tabular}}
\end{table}

\subsection{Benchmark Results and Analysis}
\label{sec:benchmark-results}

Table~\ref{tab:trajrealm-benchmark} and Fig.~\ref{fig:benchmark-radar} group methods into With GT and Without GT according to whether their tracking pipelines use GT depth and camera pose. We compare scores only within each group because scale alignment and thresholds differ. Window settings also vary by method.

Without GT geometry, strong image-space tracking does not consistently translate into accurate 3D trajectories. Track4World shows strong 2D localization on successfully evaluated sequences but weaker 3D localization than MoSE3 across the reported categories. MoSE3 outperforms TrackEverything+$\pi^3$ in 3D localization on Hard and Transportation, while TrackEverything+$\pi^3$ performs better on Embodied and Group. All three methods achieve their strongest 3D localization on Embodied, whereas sustained movement remains challenging. TrajRealm-Bench's sustained camera motion demands accurate camera parameter estimation, as estimation errors propagate into 3D trajectories and further reduce tracking accuracy.

For methods with GT depth and camera pose, Hard and Long-horizon yield the two lowest APD$_{2D}$ scores for every original model. Hard combines rapid motion, complex articulation, and frequent occlusion, while Long-horizon requires maintaining point identity through sustained movement and repeated visibility changes. TrackEverything+GT illustrates this contrast particularly clearly: its APD$_{3D}$ falls from 79.22 on Embodied to 0.44 on Long-horizon. Notably, it also performs poorly on Transportation. We found that TrackEverything is sensitive to large world-space point displacements. These results identify complex dynamics and sustained movement as major challenges even when GT geometry is available.

CoTracker3 outperforms both TAPIP3D variants in 2D and 3D localization on Hard and Long-horizon. CoTracker3 combines image-space matching with offline temporal context~\cite{karaev2025cotracker3}, whereas TAPIP3D builds local 3D feature neighborhoods and tracks through temporal windows~\cite{zhang2025tapip3d}. Rapid deformation and changing occlusion can disrupt these neighborhoods, while sustained motion requires reliable correspondence across windows. This contrast suggests that temporal association and recovery after occlusion remain critical alongside geometric modeling.

The results on Plant and Group are also interesting. Their targets, such as leaves and people, often look alike and overlap in RGB images, while occupying different depths. CoTracker3 achieves competitive 2D accuracy but substantially lower 3D accuracy than TAPIP3D. In these dynamic scenes, achieving accurate tracking across densely overlapping surfaces at different depths is difficult without using 3D information during matching. DELTA and SpatialTrackerV2 also trail TAPIP3D in 3D localization on both subsets despite using geometry during tracking. DELTA combines appearance correlations with depth differences around estimated image locations~\cite{ngo2025delta}, while SpatialTrackerV2 encodes relative 3D positions within image-grid neighborhoods~\cite{xiao2025spatialtrackerv2}. Such neighborhoods can include points from separate surfaces that overlap in the image. TAPIP3D instead selects neighbors by 3D distance and compares local neighborhoods using appearance and relative 3D positions~\cite{zhang2025tapip3d}. Its stronger performance on these subsets supports the value of using geometry to organize matching neighborhoods and distinguish targets at different depths.

\subsection{Fine-tuning and Transfer}
\label{sec:exp-results}
\label{sec:exp-controlled}

\paragraph{Comparison conventions.}
Table~\ref{tab:synthetic_tracking_comparison} evaluates TAPIP3D fine-tuning on synthetic benchmarks, while Table~\ref{tab:tracking_comparison} evaluates TAPIP3D and TrackEverything on real-world benchmarks. Unstarred rows use published baselines; starred rows report models fine-tuned on \paper. Baselines follow the respective method papers~\cite{zhang2025tapip3d,jain2026trackeverything}, except for TAPIP3D on SynthVerse, which follows SynthVerse~\cite{zhao2026synthverse}. Real-world evaluations use MegaSaM~\cite{li2025megasam} geometry for TAPIP3D and $\pi^3$~\cite{wang2026pi3} geometry for TrackEverything. The latter uses 50 full-length minival videos per dataset.

For our evaluations, AJ and OA thresholds are selected after averaging sequence scores. Avg. weights the three datasets equally and is reported when all results are available; the synthetic TAPIP3D baseline retains its published average. $\Delta$ denotes the gain over the corresponding published baseline in percentage points.

\paragraph{Transfer to synthetic environments.}
Table~\ref{tab:synthetic_tracking_comparison} shows that fine-tuning improves TAPIP3D's AJ$_{3D}$ and APD$_{3D}$ on all three synthetic benchmarks in both coordinate settings. On Dynamic Replica, AJ$_{3D}$ increases by 7.83 points in camera coordinates and 7.30 points in world coordinates. Gains also extend to LSFOdyssey and the seven-subset SynthVerse average.

\paragraph{Transfer to real-world video.}
Table~\ref{tab:tracking_comparison} shows improvements for both trackers on real-world video. TAPIP3D gains 1.19 and 1.18 points in average AJ$_{3D}$ for its camera- and world-space variants, respectively. TrackEverything improves AJ$_{3D}$ and APD$_{3D}$ on all three minival subsets, with average gains of 2.28 and 1.18 points.


%% file: sec/6_conclusion.tex
\section{Conclusion}
\label{sec:conclusion}
We introduced TrajRealm, a ten-million-frame point-tracking dataset that captures rich dynamics from games and embodied simulations. It covers object collisions, human--human and human--animal interactions, and long-distance movement through complex environments. Geometrically consistent depth and 3D point trajectories cover foreground objects and environmental surfaces. TrajRealm-Bench evaluates existing trackers across eight subcategories of Interaction and Locomotion, revealing their strengths and limitations. Experiments show that training on TrajRealm improves existing tracking methods on synthetic and real-world benchmarks. We plan to release the processed tracking data, metadata, and evaluation documentation.